\documentclass[a4paper]{article}

\usepackage[utf8]{inputenc}

\usepackage{authblk}
\usepackage{booktabs}
\usepackage[font=small,labelfont=bf]{caption}
\usepackage{graphicx}
\usepackage{natbib}
\usepackage{pdflscape}
\usepackage{url}

\newcommand*\samethanks[1][\value{footnote}]{\footnotemark[#1]}

\title{Integration of Neuromorphic AI\\
    in Event-Driven Distributed Digitized Systems:\\
    Concepts and Research Directions
}

\author{Mattias Nilsson\thanks{Embedded Intelligent Systems Lab (EISLAB),
        Luleå University of Technology,
        Luleå, Sweden.
        Email: \texttt{mattias.1.nilsson@ltu.se}
    }
    ,
    Olov Schelén\samethanks[1]
    ,
    Anders Lindgren\samethanks[1]
    \thanks{Applied AI and IoT,
        RISE Research Institutes of Sweden,
        Kista, Sweden
    }
    ,
    Ulf Bodin\samethanks[1]
    ,
    Cristina Paniagua\samethanks[1]
    ,
    Jerker Delsing\samethanks[1]
    ,
    and Fredrik Sandin\samethanks[1]
}

\begin{document}

\maketitle

\begin{abstract}
    Increasing complexity and data-generation rates in cyber-physical systems and the industrial Internet of things are calling for a corresponding increase in AI capabilities at the resource-constrained edges of the Internet.
Meanwhile, the resource requirements of digital computing and deep learning are growing exponentially, in an unsustainable manner.
One possible way to bridge this gap is the adoption of resource-efficient brain-inspired “neuromorphic” processing and sensing devices, which use event-driven, asynchronous, dynamic neurosynaptic elements with colocated memory for distributed processing and machine learning.
However, since neuromorphic systems are fundamentally different from conventional von Neumann computers and clock-driven sensor systems, several challenges are posed to large-scale adoption and integration of neuromorphic devices into the existing distributed digital--computational infrastructure.
Here, we describe the current landscape of neuromorphic computing, focusing on characteristics that pose integration challenges.
Based on this analysis, we propose a microservice-based framework for neuromorphic systems integration, consisting of a neuromorphic-system proxy, which provides virtualization and communication capabilities required in distributed systems of systems, in combination with a declarative programming approach offering engineering-process abstraction.
We also present concepts that could serve as a basis for the realization of this framework, and identify directions for further research required to enable large-scale system integration of neuromorphic devices. \end{abstract}

\section{Introduction}
\label{sec:introduction}

The accelerating developments of digital computing technology and deep learning--based AI are leading towards technological, environmental, and economic impasses \citep{thompson2021deep, mehonic2022brain}.
With the end of Dennard transistor-scaling \citep{davari1995cmos} and the anticipated end of Moore’s law \citep{waldrop2016chips, shalf2020future, leiserson2020there}, conventional digital von Neumann computers and clock-driven sensor systems face considerable hurdles regarding bandwidth and computational efficiency. For example, the gap between the computational requirements for training state-of-the-art deep learning models and the capacity of the underlying hardware has grown exponentially during the last decade \citep{mehonic2022brain}.
Meanwhile, in stark contrast, distributed digitized systems---ever-growing in size and complexity---require increasing computational efficiency for AI applications at the resource-constrained edge of the internet \citep{zhou2019edge, ye2021challenges}, where sensors are generating increasingly unmanageable amounts of data.

One approach to addressing this lack of computational capacity and efficiency is offered by \textit{neuromorphic engineering} \citep{mead1990neuromorphic, mead2020we}.
There, inspiration is drawn from the most efficient information processing systems known to humanity---brains---for the design of hardware systems for sensing \citep{tayarani2021event} and processing \citep{zhang2020neuro, basu2022spiking} that have the potential to drive the next wave of computational technology and artificial intelligence \citep{christensen2022roadmap, frenkel2021bottom, shrestha2022survey}.
Neuromorphic---that is, brain-like---computing systems imitate the brain at the level of organizational principles \citep{indiveri2015memory}, and often also at the level of device physics by leveraging nonlinear phenomena in semiconductors \citep{chicca2014neuromorphic, rubino2021ultra} and other nanoscale devices \citep{zidan2018future, markovic2020physics} for non-digital computation.
The idea of using nonlinear physical phenomena for non-digital computing has been explored for decades.
Different choices of underlying mathematical models lead to different definitions of what the concept of “computation” entails \citep{jaeger2021toward}, and likely also influences the set of possible emergent innovations.

Here, we define neuromorphic computing (NC) systems as non--von Neumann information-processing systems, the structure and function of which either emulate or simulate the neuronal dynamics of brains---especially of somas, but sometimes also synapses, dendrites, and axons---typically in the form of \textit{spiking neural networks (SNNs)} \citep{maass1997networks, nunes2022spiking, wang2022hierarchical}.
NC systems open up new algorithmic spaces---through asynchronous massive parallelism, sparse, event-driven activity, and co-location of memory and processing \citep{indiveri2015memory}---and, in terms of energy-usage and latency, offer superior solutions to a range of brain-like computational problems \citep{davies2021advancing, yin2021accurate, stockl2021optimized, goltz2021fast, rao2022long}.
Furthermore, beyond cognitive applications, SNNs and NC systems have also demonstrated potential for applications such as
graph algorithms,
constrained optimization,
random walks,
partial-differential-equation solving,
signal processing,
and algorithm composition \citep{aimone2022review}.
Consequently, there is a growing interest for NC technology within application domains such as
automotive technology,
digitized industrial production and monitoring,
mobile devices,
robotics,
biosensing (such as brain--machine interfaces and wearables),
prosthetics,
telecommunications-network (5G/6G) optimization,
and space technology.

One challenge facing neuromorphic technology is that of integrating emerging diverse hardware systems, such as neuromorphic processors and quantum computers, into a common computational environment \citep{vetter2018extreme}.
Such hardware systems are---due to performance constraints of existing computational hardware in, for instance, energy usage or processing speed---likely to be increasingly included in computational ecosystems to facilitate or accelerate particular types of computation \citep{shalf2020future, leiserson2020there, hamilton2020accelerating}.
Fundamental trends in computer-architecture development indicate that nearly all aspects of future high-performance computing architectures will have substantially higher numbers of diverse and unconventional components than past architectures \citep{becker2022unconventional}, leading toward a period of “extreme heterogeneity”.
Consequently, neuromorphic processors are, in many future use-cases, likely to be part of a broader, heterogeneous computational environment, rather than to be operated in isolation.
Thus, there is a need for programming models and abstractions, as well as interparadigmatic communication principles and data models, that enable effective integration of neuromorphic hardware into large-scale systems of systems.

Here, we address the problem of large-scale adoption and integration of NC systems into the present digital--computational infrastructure.
We frame this problem in terms of the following main challenges:
\begin{enumerate}
    \item \textbf{Communication:}
    How to transcode information between neuromorphic and digital systems?
    
    \item \textbf{Virtualization:}
    How to interface neuromorphic devices and services in distributed digitized systems?
    
    \item \textbf{Programming:}
    How to efficiently program hybrid neuromorphic–digital systems?

    \item \textbf{Testing and validation:}
    How to reliably train and test the functionality of such hybrid systems?
\end{enumerate}

We outline the current landscape of NC technology from the perspective of system integration---describing the most significant qualities of NC systems as compared to the fundamentally different paradigm of conventional digital computing (DC)\footnotemark.
Based on this description, we propose a microservice-based framework for integration of NC systems.
The framework consists of a neuromorphic-system proxy, which provides virtualization and communication capabilities required in a distributed setting, in combination with a declarative programming approach offering engineering-process abstraction.
We present established concepts for programming, representation, and communication in distributed systems that could serve as a basis for the realization of this framework, and identify directions for further research required to enable such integration.

\footnotetext{The terms “digital computing (DC) system” and “von Neumann computer” are used interchangeably throughout this article.
}
 
\section{Neuromorphic Systems}
\label{sec:nc_systems}

The field of neuromorphic engineering dates back to the late 1980s \citep{mead1990neuromorphic, mead2020we}, and originally dealt with the creation and use of sensing and processing systems that imitate the brain at the level of structure and device physics.
Today, the term “neuromorphic” has broadened, and “neuromorphic processors” typically refer to hardware systems of different architectures that are specialized for running SNNs.
Neuromorphic hardware architectures thus range from electronic emulation with analog circuitry or novel electronic devices to digital systems specialized for massively parallel differential-equation solving for spiking neuron models.
However, as spiking neural networks \citep{maass1997networks, nunes2022spiking, wang2022hierarchical} are inherently event-driven, asynchronous, time-dependent, and highly parallel, all neuromorphic processors, by consequence, differ significantly from von Neumann computers, as summarized in \textbf{Table~\ref{tab:architectures}}.
\begin{landscape}
    \begin{table}[tb]
\centering
        \caption{\textbf{Qualitative differences between von Neumann and neuromorphic computational architectures.}
        }
        \begin{tabular}{l l l}
            \toprule
            \textbf{Architecture}                       & von Neumann                   & Neuromorphic \\
            \midrule
            \textbf{Processing operations}              & Sequential                    & Massively parallel \\
            \textbf{Memory--processing organization}    & Centralized, separated        & Distributed, colocated \\
            \textbf{Temporal organization}              & Synchronous, clock-driven     & Asynchronous, event-driven \\
            \textbf{State qualities}                    & Discrete, static              & Continuous, dynamic \\
\textbf{Programming method}                 & Sequential logic              & Structural SNN configuration \\
            \textbf{Unit of communication}              & Binary numbers                & Unary spike-events (spatiotemporal, sparse) \\
            \bottomrule
        \end{tabular}
        \label{tab:architectures}
    \end{table}
\end{landscape}
In general, analog-based NC systems are more power-efficient than fully digital ones \citep{basu2022spiking}, by leveraging device physics for real-time neurosynaptic emulation, while digital systems come with the versatility of being fully configurable by logical programming.
Due to the need for power-efficient sustainable technologies for AI workloads, neuromorphic solutions can come to constitute up to 20~\% of AI computing and sensing by 2035 according to some estimates.

\subsection{States in Neuromorphic Systems}
\label{sec:nc_states}

The state of a neuromorphic processor at any given moment is defined by the properties of the SNN that that processor has been configured to implement and the event-based neurosynaptic activity of that SNN, which is largely reactive in response to input signals.
These properties can roughly be arranged into the following categories:
\begin{itemize}
    \item \textbf{Structural properties:}
    E.g., network topology, synaptic weights, time constants, axonal delays, and neuronal thresholds
    
    \item \textbf{Transient properties:}
    E.g., neuronal potentials, synaptic currents, and spiking activity
\end{itemize}
Out of these properties, it is, in general, the structural ones that are subject to direct manipulation by external configuration, optimization, and learning algorithms---while the transient state rather arises in reaction to presented input signals in concert with the structural state.
However, a clear line cannot simply be drawn between structural and transient properties, as the biological timescales of synaptic plasticity phenomena---that is, the strengthening and weakening of neuronal connections---range from single milliseconds, in the case of short-term plasticity, to the whole lifetime of an organism, in the case of structural plasticity \citep{jaeger2021dimensions}.
Many neuromorphic systems do, however, exclude on-chip implementation of synaptic plasticity due to the complexity and resource cost \citep{frenkel2021bottom}, in which case, structural parameters are more clearly distinguished as subject to configuration by an external system.
There are many learning rules in use due to the knowledge gap associated with long-term plasticity and task-dependent requirements.
Therefore, some chips implement flexible coprocessors for learning.

\subsection{Information in Neuromorphic Systems}
\label{sec:nc_information}

Von Neumann computers represent information in clock-driven discrete states, the resolutions of which are determined by the number of bits used for representing binarily encoded variables.
NC systems, on the other hand, represent information using unary (one-or-nothing), uniform interneuronal spike-events.
These carry explicit information about both space and time in their source of origin and time of arrival---potentially carrying arbitrary temporal precision in the interspike intervals \citep{thorpe2001spike}.
This form of representation arises already in neuromorphic sensors, as they rely on level-crossing Lebesgue sampling \citep{astrom2002comparison} for event-driven generation of sense data, or, alternatively, in delta-modulated spike-data conversion of conventionally sampled signals \citep{corradi2015neuromorphic}.
Spike-based representations thus allow sparse, event-driven processing with consequently low power-usage---especially for real-time applications in which both sensing and processing are spike-based and event-driven \citep{liu2019event}.

\subsubsection{Neural Code}

There are several ways in which spatiotemporal combinations of uniform spikes could theoretically be used to encode information.
\cite{thorpe2001spike} outline the following theoretical spike-based neural coding schemes:
\begin{enumerate}
    \item \textbf{Rate code:}
    Information is represented by \textit{how often} each single neuron fires, in the form of a time-averaged firing rate.
    (ANNs are based on rate code.)
    
    \item \textbf{Count code:}
    Information is represented by \textit{how often} a group of neurons fire in total, during a temporal interval.
    
    \item \textbf{Timing code:}
    Information is represented by \textit{when and where} each spike occurs---that is, in the source of origin and time of arrival.
    
    \item \textbf{Rank-order code:}
    Information is represented by the \textit{temporal order} in which a group of neurons fire, but without further spike-timing information.
    
    \item \textbf{Synchrony code:}
    Information is represented by \textit{which} neurons fire closely in time to each other during a temporal interval.
\end{enumerate}
This list is not exhaustive, and the manner in which information is actually represented in the brain is still largely an open question \citep{brette2015philosophy, zenke2021visualizing}.
It is, for instance, possible that the asynchronous dynamics of SNNs give rise to emergent representations that consist of combinations of coding schemes such as those listed above.
Nevertheless, the listed coding schemes---along with their estimated capacity for information transmission, see \textbf{Table~\ref{tab:coding_schemes}}---provide an overview of the qualitatively distinct ways in which information can be represented in SNNs, and the quantitative relations between these.

\begin{landscape}
    \begin{table}[tb]
\caption{\textbf{Information-transmission capacity of different neural coding schemes.}
Adapted from \cite{nilsson2021using}.
Arranged in descending order of capacity, as theoretically estimated by \cite{thorpe2001spike}.
\textit{N} is the number of neurons, each spiking maximally once.
The number of equivalent bits were calculated for the example scenario of \textit{N}~=~10 and a temporal interval of \textit{t}~=~10~ms.
        }
        \centering
        \begin{tabular}{l c c l}
            \toprule
            \textbf{Coding scheme}  & \textbf{Possible states}  & \textbf{Equivalent bits}  & \textbf{Comments} \\
                                    & \textbf{(no.)}            & \textbf{(no.)} \\
            \midrule
            Timing  code    & $(t/\delta t)^{N}$    &   33      & For temporal resolution $\delta t$ = 1~ms \\
            Rank-order code & $N!$                  &   21      & Temporal ordering \\
            Synchrony code  & $n_{\Phi}^{N}$        &   20      & For $n_{\Phi} = 3$ possible phases\\
            Binary code     & $2^{N}$               &   10      & Used in von Neumann computers \\
            Count code      & $N + 1$               &   3.46    & Equivalent to rate code in this scenario \\
            \bottomrule
        \end{tabular}
        \label{tab:coding_schemes}
    \end{table}
\end{landscape}

The estimates presented in \textbf{Table~\ref{tab:coding_schemes}} were made for a population of \textit{N} = 10 neurons, during a temporal interval of \textit{t} = 10 ms, and with a temporal resolution of 1 ms.
As the estimations were made for scenarios of rapid processing, they were also limited to a maximum of one spike per neuron.
During these conditions, rate code is theoretically equivalent to count code, as rate code would require more than a single spike per neuron---and thus a longer duration---to represent more information.
While these estimates were made for limited conditions, it illustrates how the information transmission per spike would be maximal if spikes carried information in their precise timings.
However, a spike-timing-based coding scheme may demand a high level of complexity and temporal precision in the decoding mechanisms.
It is important to note here that the data generated by neuromorphic, event-driven sensors is, at least in part, intrinsically spike-timing coded due to the event-driven, sparse activations that underlie the low-power, low-latency operation of such sensors \citep{liu2019event}.
Therefore, in order to gain analogous benefits in the subsequent processing, it is likely necessary to incorporate some degree of spike-timing code in neuromorphic processing systems.
However, the choice of coding scheme is likely to be task-specific and subject to optimization \citep{guo2021neural, schuman2022evaluating}.

\subsubsection{Representation Space}

\begin{figure}[tb]
    \centering
    \includegraphics[width=0.7\textwidth]{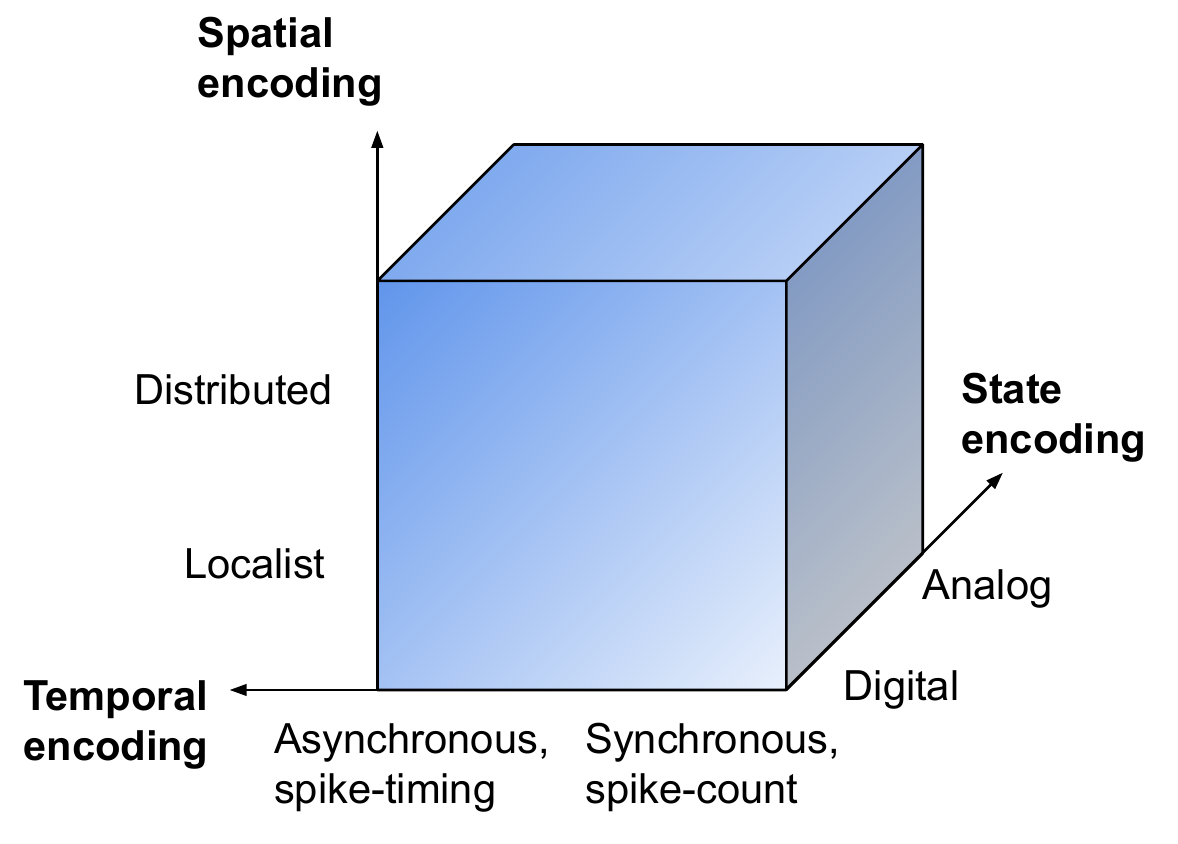}
    \caption{\textbf{Space of possible representations in neuromorphic systems.}
The illustration depicts a continuum, in which gradual, independent changes along each dimension are possible, as there exists many variations and combinations of the mentioned concepts.
}
    \label{fig:rep_space}
\end{figure}

\textbf{Figure~\ref{fig:rep_space}} illustrates a conceptualization of the space of possible information representations in NC systems.
As discussed previously, the NC hardware substrate may, to varying degrees, rely on digital or analog circuitry, and the temporal encoding may, again to varying degrees, asynchronously rely on the precise timings of spikes in qualitatively different coding schemes, see \textbf{Table~\ref{tab:coding_schemes}}.
The spatial dimension of a neural network is generally, by default, used for \textit{distributed representations}, in which the representations of different concepts are distributed over several of the same neurons and synapses of the network.
Conversely, \textit{localist representations}, which are studied in conventional, logical neurosymbolic computation \citep{garcez2020neurosymbolic, dold2022neuro}, represent different concepts with single, discrete identifiers, such as single neurons or bits.
An example of a completely localist representation could, for instance, be a single “cat neuron”, which, when activated, signifies the inferred presence of a cat in the sense data.
As in the case of the two other dimensions of \textbf{Figure~\ref{fig:rep_space}}, it also depicts a possible spectrum of spatial encoding, in the hypothetical extremes of which, representations are either distributed across a whole neural network or localized to single neurons, respectively.

\subsubsection{Interfaces and Semantics}
\label{sec:nc_interfaces}

For the activities of a neural network to ultimately have discernible implications for action in an external system, \textit{symbolic representations} are needed---such as, for example: Pattern~A implies Action~X and Pattern~B implies Action~Y.
“Action” is here meant in a broad sense, ranging from motor-output actuation to read-out signaling.
In the kind of hybrid digital--neuromorphic systems discussed here, such symbolic representations would, depending on the application, not necessarily have to be generated within the confines of the NC system itself.
However, at every system interface, there needs to be an operational \textit{semantic}---a system of interpretation---for transmitted data \citep{JNilsson2022PhDThesis,nilsson2018semantic}.
Such interfaces could, for instance, be implemented in the form of microservices, which is further discussed in Section~\ref{sec:framework}.

An interesting concept for interface semantics for NC systems is that of \textit{embeddings}---a widely used form of representation in neural networks and machine learning that are essential to many state-of-the-art models.
Embeddings are succinct, intermediate representations of associations within usually large-scale datasets.
For example, embeddings are used in deep learning architectures where concept representations are quickly learned from combinations of sensor data, descriptive sentences, and higher-level knowledge representations \citep{mei2022falcon}.
The learned concepts can be used in downstream applications, such as answering questions by reasoning about unseen sensor inputs.
Typically, embeddings emerge as nested function values in a deep learning model,
$y_i = f_n(\ldots f_0(x_i; \theta_0);\theta_n)$.
The parameters $\theta_k$ are optimized to provide the outputs $y_i$ that are expected for the corresponding inputs $x_i$ with minimum error according to some metric.
The embeddings, $f_k(\ldots)$, of a model that is optimized on a sufficiently large and varied dataset are often useful for other similar datasets, as these embeddings can be used as input to optimize a new output function with less data and computational resources compared to full model retraining.
Embedding are used in various models---autoencoders, transcoders, multimodal predictive models, etc.---and are natural components to be used and supported by the envisioned hybrid neuromorphic--digital computing systems and programming models.
This entails interesting opportunities and challenges related to the optimization and interoperability of embeddings realized in different hardware and all the related symbols appearing in software and data across an orchestrated system \citep{nilsson2020interoperability,nilsson2019interoperability}.

\subsection{Programming of Neuromorphic Systems}
\label{sec:nc_programming}

In contrast to von Neumann computers, the actions of NC systems are not primarily dictated by sequences of explicit logical instructions.
Rather, the analog to DC programs in NC systems can be considered as being implicitly defined by the structural properties---as defined in Section~\ref{sec:nc_states}---of the implemented SNNs, as these determine the input--output signal transformations that are performed by an NC system.
Furthermore, while von Neumann computers encode information in observable, static, discrete states, NC systems may inhabit unobservable, dynamic, continuous states, and they receive inputs in the form of uniform spike-events, in which information is encoded in the physical time of arrival and source of origin.
Thus, the use of NC systems is fundamentally different from that of von Neumann computers, and---in order to fulfil the potential for efficiency---may require a significant change of perspective in the view of programming \citep{schuman2022opportunities} and computation \citep{jaeger2021dimensions}, informed by neuroscience and dynamical-systems theory.

Some attempts to develop programming abstractions for NC systems include the Neural Engineering Framework (NEF) \citep{stewart2012technical} and Dynamic Neural Fields (DNFs) \citep{sandamirskaya2014dynamic}.
However, these are often fairly limited to specific use-cases---biologically plausible neural models for the NEF, and models of embodied cognition for DNFs.
Thus, there is a gap in defining more generally useful programming abstractions for neuromorphic computing systems \citep{schuman2022opportunities}, including virtualization concepts required for seamless edge-to-cloud integration.

\subsubsection{Software}
\label{sec:nc_software}

As the landscape of neuromorphic computing is made up of different hardware and software architectures that are developed by different groups, it is characterized by a fragmented and noncomposable array of programming models and frameworks.
Programming frameworks for SNNs and neuromorphic hardware generally fall into one of the following categories:
\begin{itemize}    
    \item \textbf{Optimization tools:}
    SNN-parameter optimization tools, usually based on supervised deep learning, such as
    SNN Conversion Toolbox \citep{rueckauer2017conversion},
    SLAYER \citep{shrestha2018slayer},
    Whetstone \citep{severa2019whetstone},
    EONS \citep{schuman2020evolutionary},
    and EXODUS \citep{bauer2022exodus}
    
    \item \textbf{Simulators:}
    SNN simulators with low-level APIs for conventional computers, such as
    NEST \citep{gewaltig2007nest},
    Brian~2 \citep{stimberg2019brian2},
    Nengo \citep{bekolay2014nengo}
    GeNN \citep{yavuz2016genn},
    BindsNET \citep{hazan2018bindsnet},
    SynSense Rockpool and SINABS, and Norse 

    \item \textbf{Hardware interfaces:}
    Low-level interfaces and runtime frameworks for configuration of neuromorphic hardware, such as
    PyNN \citep{davison2009pynn},
    Fugu \citep{aimone2019fugu},
    SynSense Samna, BrainScaleS OS \citep{muller2022brainscalesos},
    and Intel Lava
\end{itemize}
While several frameworks exist, none of them have so far provided programming abstractions that are composable and span the diverse range of algorithms and methods within NC \citep{davies2021advancing, jaeger2021toward}.
Furthermore, NC hardware typically have limitations in terms of connectivity, plasticity, and neurosynaptic configurations.
Thus, transforming a well-defined SNN specification or general program into a corresponding hardware configuration is challenging and further complicated by imperfections of mixed-signal circuits and limited resources, such as bandwidth, which generate differences from the specified target.
Even between generations of the same hardware architecture, such as Spikey and BrainScaleS-1, SpiNNaker-1 and -2, or Loihi-1 and -2, it is often difficult to build upon existing software \citep{muller2022brainscalesos}.
As of today, two candidate models for general-purpose and platform-agnostic NC configuration are PyNN and Lava.

\paragraph{PyNN}

PyNN \citep{davison2009pynn} is a simulator-agnostic language for describing SNN models at the level of network topology, neurosynaptic parameters, plasticity rules, input stimuli, and recording of states, while still allowing access to the details of individual neurons and synapses.
PyNN also provides a set of commonly used connectivity algorithms (e.g., all-to-all, random, distance-dependent, small-world) but makes it easy to provide custom connectivity in a simulator-independent way.
PyNN provides a library of standard models of neurons, synapses, and synaptic plasticity, which have been verified to work the same on the different supported simulators.
As of today, common SNN simulators and some hardware emulators support PyNN, which is also the entry-point to the BrainScaleS and SpiNNaker systems that implement PyNN as an experiment-description language.

\paragraph{Lava}

Lava\footnotemark\ by Intel is an upcoming open-source software framework for neuro-inspired applications and their deployment on neuromorphic hardware, and constitutes an attempt to move towards convergence in the domain of NC software.
\footnotetext{https://lava-nc.org/}
Lava is designed to be hardware-agnostic, modular, composable, and extensible---allowing developers to construct abstraction layers to meet their needs, and to broaden the accessibility of programming NC systems.
The fundamental building-block in Lava, for algorithms and applications alike, are so-called \textit{processes}---stateful objects with internal variables and input and output ports for message-based communication via \textit{channels}.
This architecture is inspired by the communicating sequential processes (CSP) formal language for asynchronous, parallel systems, which belongs to the family of formal models for concurrent systems known as \textit{process calculus}, further described in Section~\ref{sec:declarative}.
Every entity in Lava---including neurons, neural networks, conventional computer programs, interfaces to sensors and actuators, and bridges to other software frameworks---is a process with its own memory and message-based communication with its environment.
Thus, Lava processes are recursive programming abstractions, from which, modular, large-scale parallel applications can be built.

\subsection{Challenges to Adoption and Integration}

The following are some of the major challenges posed to the adoption of NC technology and its integration into the present computational environment.

\subsubsection{Programming Abstractions and Frameworks}

As discussed in Section~\ref{sec:nc_programming}, there is a lack of common programming abstractions, models, and frameworks for different NC designs  \citep{davies2021advancing, schuman2022opportunities}.
Intel's launch of the Lava software framework is an attempt at closing this gap, but, being so recent, the degree to which Lava will aid in achieving NC software convergence remains to be proven.
Furthermore, there may be a need for further developments of generalized system hierarchies, concepts of completeness \citep{zhang2020system}, and analytical frameworks \citep{guo2021marr} for NC systems and other unconventional computing concepts \citep{jaeger2021toward}, to facilitate hardware--software compatibility, programming flexibility, and development productivity.

\subsubsection{Interdevice Communication}

Most neuromorphic systems---sensors and processors alike---implement an address-event representation (AER) spike-event communication protocol \citep{mortara1994communication, boahen2000point}, in which events, signified by the address of their source of origin, such as a neuron or pixel, are asynchronously generated and transmitted in real-time along the connections of neural networks.
However, although it is standard practice to implement \textit{some} AER protocol, there are slight differences in the implementations between the different neuromorphic sensory and processing devices that currently exist \citep{basu2022spiking}, as these are developed by different groups.
This discrepancy between different neuromorphic devices impairs their interoperability, as well as standardization of NC--DC communication, and thus poses a challenge to the integration of neuromorphic systems into the broader computational environment.

\subsubsection{Reliance on Host-Computers}

Currently, the use of NC systems relies heavily on conventional host-computers for software deployment and, often, for communication with the environment via sensors and actuators.
This reliance on a host-machine---which performs preparation and deployment of the NC model and pre- and post-processing of spike-data---can impact the resource requirements for running the neuromorphic system to such an extent that the performance benefits of using such specialized hardware are lost \citep{diamond2016comparing}.
Thus, there is a need to optimize the host--device communication architecture with regard to scalability, throughput, and latency, as well as to design and implement SNNs and NC systems in a way that minimizes the need for host--device communication in the first place. 
\section{Integration Framework}
\label{sec:framework}

\begin{figure}[tb]
    \centering
    \includegraphics[width=0.9\textwidth]{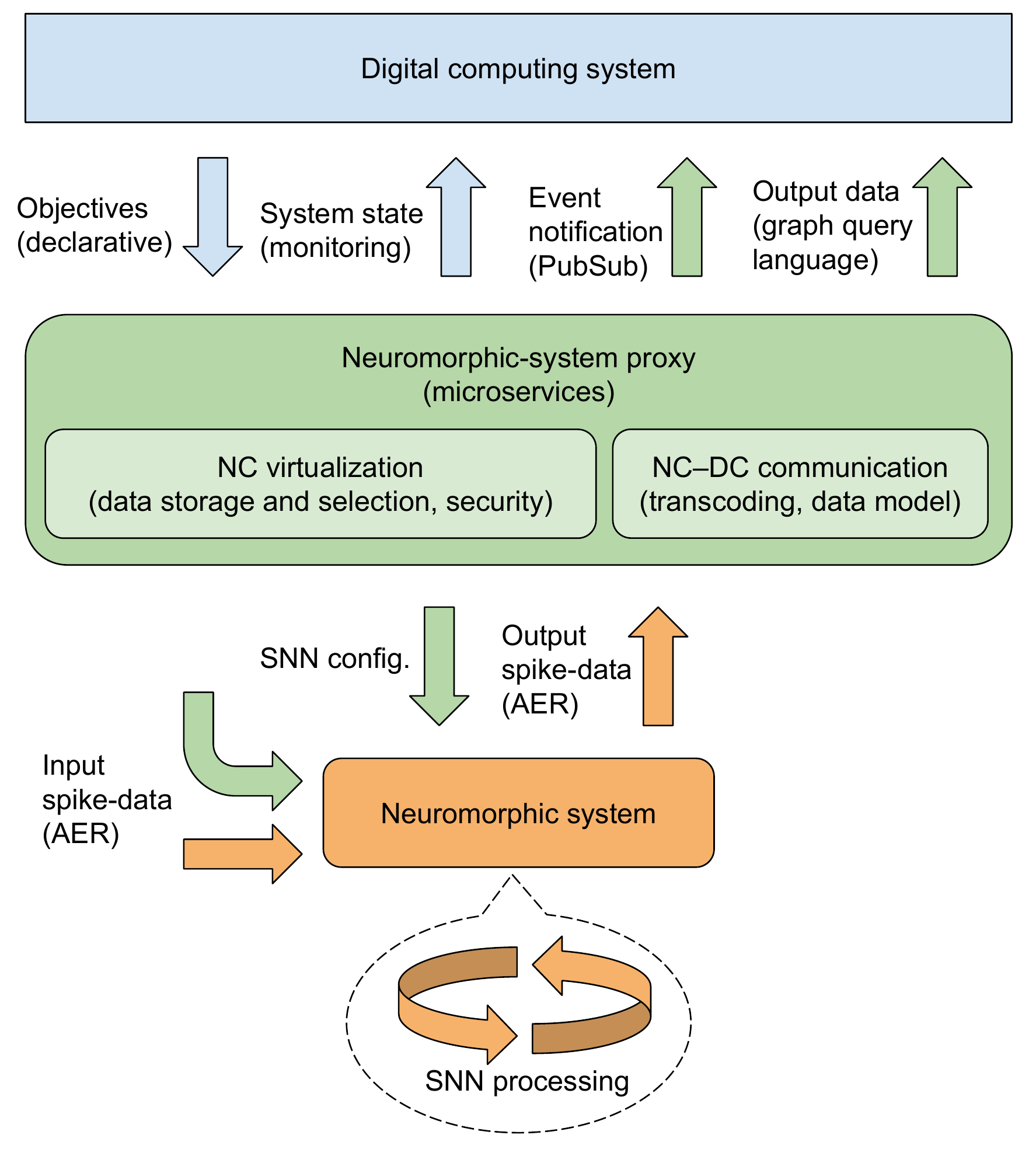}
    \caption{\textbf{System-integration framework for neuromorphic devices in digital--computational infrastructure.}
Arrows symbolize information channels.
The coloring indicates information being represented and transmitted within the following domains:
        orange for spike-based neuromorphic sensing and processing,
        green for a virtual neuromorphic-system proxy,
        and blue for conventional digital computing.
    }
    \label{fig:integration_concept}
\end{figure}

In this section, we outline a framework for the integration of NC systems into the existing environment of distributed DC systems.
The framework, see \textbf{Figure~\ref{fig:integration_concept}}, consists of NC--DC abstraction layers and communication channels that we envision to be necessary for the NC-system integration. 
The framework is aligned with the paradigm of \textit{microservices} \citep{larrucea2018microservices}---small, single-responsibility applications, inspired by service-oriented computing, that can be deployed, scaled, and tested independently---and adapted to the neuromorphic context so that there is a separation between the digital and neuromorphic parts.
Specifically, the digital abstraction is provided by a microservice instance, called a neuromorphic-system proxy (NSP), illustrated in the middle box of \textbf{Figure~\ref{fig:integration_concept}}.
It represents an NC system, see the bottom box in \textbf{Figure~\ref{fig:integration_concept}}.
The role of the NSP is to provide virtualization and data mapping between the neuromorphic and digital domains, and thereby providing effective and efficient interfacing between the digital and neuromorphic domains, including for example availability and security.

The use of microservice architecture provides characteristics such as late binding, loose coupling, and discovery to the framework.
In addition to these characteristics, a major benefit of the use of microservices in the NSP is service longevity.
Service longevity implies that services are available over time.
In contrast to the NC systems that provide transient data in real-time reaction to events, the NSP stores the last data communicated from the NC system and exposes it via the available microservices.

The NSP is software-defined and typically built on a library of base-software images (e.g., docker images) stored in a repository to provide easy instantiation and replication at the edge or cloud as desired.
Application-specific programs are added as delta images.
Although the figure shows a one-to-one mapping between an NC system and an NSP, this model can be scaled out to arbitrary numbers of NC systems and NC system proxies.

The NSP is here described by first covering the essential interfaces, i.e., the services provided or consumed and then, in Section~\ref{sec:data_models}, some challenges and approaches for data modeling and mapping are described.
The interfaces to an NSP are represented by arrows in \textbf{Figure~\ref{fig:integration_concept}} as follows:

Bottom part (related to NC-system interfaces):
\begin{itemize}
    
    \item \textbf{SNN config.:}
Drives the SNN configuration as needed for specific use cases.
This is slowly changing data.
Initially, we envision the configuration to be primarily static, while it may be somewhat dynamic to allow some degree of digital reconfiguration as part of normal runtime operation.

    \item \textbf{Input spike-data:}
The input to the SNN, in which most spikes typically come from other neuromorphic systems such as sensors.
In addition, stimuli from other systems can also enter through the digital interfaces of the NSP.
All incoming spikes are processed by the NC system with the objective to produce an output that aligns with the requirements expressed by the declarative definitions and the validation protocol.

    \item \textbf{Output spike-data:}
Involves information from selected spike-sources, such as read-out neurons (as defined by the SNN config.).
Typically, abstractions are made such that only spikes carrying high-level information are communicated to the NSP.
This serves to reduce the load of events coming to the NSP, although the transmitted data is still highly dynamic.
    
\end{itemize}

Top part (related to digital interfaces):
\begin{itemize}

    \item \textbf{Objectives:}
Interface for adding declarative statements that will be translated to an SNN config. 
This is complementary to code that is part of the NSP image.
The need depends on the dynamics of reconfiguration required, and using it is therefore optional.

    \item \textbf{System state:}
Interface for retrieving the SNN configuration in terms of structural SNN properties, see Section~\ref{sec:nc_states}.

    \item \textbf{Event notification:}
Interface where events are sent from the NSP.
In alignment with the microservice paradigm, these events are typically sent to some PubSub event-handlers using different channels or topic trees to notify subscribers effectively and efficiently.

    \item \textbf{Output data:}
Interface where other systems can request additional data related to the events seen in the event notification or related to any history or state stored in the NSP.
This is basically a restful interface, but aiming for a graph query language where there is a well-defined typed data schema and where under- or over-fetching is avoided.

\end{itemize}

Within the NSP, a virtualized representation of the neuromorphic system (a digital twin proxy) provides key functionality to allow for representations of the NC system to exist on one or more digital systems, maintaining the state of the system.
The virtualization component also provides the essential functionality of data selection and storage that makes it possible to aggregate and select which data from the NC system should be exposed to the digital part of the system.
This component also provides storage, allowing access to data \textit{a posteriori} to its creation--- either through short-term caching, which allows efficient access for multiple clients, or longer-term storage for historical access to data.

In the subsequent parts of this section, we discuss established computer-scientific concepts that could be used to realize different parts of this framework or, if insufficient, that could inform the extension or development of new such concepts.

\subsection{Communication and Data Models}
\label{sec:data_models}

In order to integrate NC systems with other entities in distributed DC systems, the standardized protocols and paradigms used for information representation and communication in the DC domain need to be interfaced within the NSPs.
This is a challenging problem considering the different information representations used in the DC and NC domains, including for example state-based and time-based representations, differences in semantics, and a variety of information encoding and transcoding methods used.
Here, we review some well-established DC concepts that share characteristics with the NC domain, and may thus be helpful to address the challenges of data modeling and information translation and representation.

\subsubsection{Soft-State Models}

The event-based nature of NC systems maps well to the best-effort message-passing based nature of many current computer networks, such as the Internet. 
A key feature of best-effort computer networks to provide a distributed system state across the network, in a scalable manner that is robust to network disruptions and do not require too much communication overhead, is the concept of \textit{soft state}~\citep{Ji2007}.
This is used in Internet routing protocols and other networked distributed systems and becomes even more important in more unstable systems such as \textit{ad hoc} or delay tolerant networks~\citep{Perkins2003, Fall2008, Grasic2011}.

In a soft-state system, a distributed state is maintained in individual nodes by periodic update messages (or \textit{events}) that are sent by other nodes in the network to refresh the current state.
Such state can for example be used to maintain routes through the network for forwarding data traffic. 
If no refresh messages have been received within a certain amount of time, it is assumed that the state of the system has changed (e.g., a neighboring node has moved away and can no longer be reached), and thus the local state is removed.
Since the underlying network is best-effort and refresh messages can be lost due to contention or interference, the system might wait until more than one expected refresh message has not been received before determining that the local state should be removed.
The threshold used to determine a state change depends on the expected characteristics of the network and is a tradeoff between how rapidly the system reacts to state changes and the desired robustness to temporary disruptions.
The benefit of utilizing soft state is that there is no risk of stale state information becoming stored indefinitely in parts of the network that cannot be reached by state-removal messages, and the need for such control traffic is removed completely.

The soft-state model is similar to an NC system, where messages (spike-events) trigger and maintain state in nodes (neurons).
Thus, we believe that there are potential benefits from mapping these models in the NC domain to digital integration in terms of data models.

\subsubsection{Event Notification}

The Internet has enabled the creation of very large distributed systems, commonly with the purpose of sharing information between geographically distributed data producers and consumers.
In such systems, the publish--subscribe (PubSub) messaging paradigm, based on event notification and messaging, has received much attention for the loosely coupled form of interaction it provides in large-scale settings \citep{liu2003survey}.
In this paradigm, subscribers register their interests in a topic or a pattern of events and then asynchronously receive events matching their interest.
PubSub provides decoupling in space (interacting subscribing and publishing parties do not need to know each other), time (parties do not need to interact at the same time), and asynchronous communication (subscribers can get notifications immediately when published) \citep{10.1145/857076.857078}.

PubSub systems can be categorized into subject-based, aka topic-based, and content-based \citep{liu2003survey}.
Subject-based PubSub means that a subscription targets a group, channel, or topic, and the user receives all events that are associated with that group.
With content-based PubSub systems, subscriptions are instead based on queries or predicates, based on which, the decision of to whom a message is directed is made on a message-by-message basis.
The advantage of a content-based system is that the subscriber can be provided with the needed information only and does not need to learn a set of topic names and their content before subscribing.

The performance of a content-based PubSub network is typically challenged by the expensive matching cost of content messages.
Hybrid schemes between these types of PubSub systems exist, where subscribers register content-based subscriptions to one or more topics.
For example, HYPER minimizes both the number of matches inside the PubSub network and the delay to receive subscribed content \citep{1437105}.
The hybrid schemes, aka type-based, thus represent the middle-ground between coarse-grained, topic-based systems and fine-grained, content-based systems.
Type-based systems gives a coarse-grained structure on events (like in topic-based) on which fine-grained constraints can be expressed over attributes (like in content-based) \citep{Shen2010}.

PubSub systems based on event notification and messaging have been adopted by the Service-Oriented Architecture (SOA) model \citep{1210138}\citep{5072495} and its later incarnation known as the Application Program Interface (API) and microservice paradigm \citep{7958492}.
In contrast to traditional SOA, the API and microservice paradigm build on the requirement of that such service must be independently deployable \citep{7880473}.
Modern messaging systems are typically based on message-oriented middleware like Apache Kafka, Rabbit MQ, NATS Streaming, Google Pub/Sub, Microsoft Event Hubs and Amazon Kinesis \citep{10.1145/857076.857078}.

We believe that the properties provided by PubSub and event notification as used in SOA and microservice-based systems, i.e., space, time and synchronization decoupling, together with the ability to efficiently monitor states and outcome of NC systems, can prove useful to make NC systems interact efficiently with DC systems.
For example, event notification can be used to learn about that something has happened in an NC system that motivates further investigation based on more output data requested using a graph query language like GraphQL, and tools for its use like Apollo GraphQL and Graphene Python.

\subsubsection{Graph Data and Querying}

Graph theory is one of the corner stones of computer science.
A simple graph $G$ \citep{wilson10} consists of a non-empty finite set $V(G)$ of elements called vertices (or nodes), and a finite set $E(G)$ of distinct unordered pairs of distinct elements of $V(G)$ called edges.
A generalization of this is the multigraph, which can have multiple edges between the same pairs of vertices.
Often, the edges are labeled to represent different kinds of relationships between vertices. 
Other generalizations of graphs include weighted graphs where the edges have a numeric weight, and directed graphs where each edge has a direction.
In graph databases, it is common to represent directed multigraphs, where both vertices and edges are labeled and annotated with additional data such as weights and properties.
This representation is general and can be used to model all abovementioned kinds of graphs.
We believe that the graph is a strong candidate to model both neuromorphic structures and the structure of abstract data that neuromorphic sensors can produce---at least to a large extent.

Real life graphs can be complex and have large sizes.
In general, they can include cycles, unless we have defined the special case of a directed acyclic graph (DAG).
To understand properties of graphs, it is common to traverse them to search for paths or data of interest, where traversals must terminate and not loop cycles indefinitely.
It should be noted that the general problem of finding an optimal path through a graph is NP hard.
There are many alternative traversal algorithms available, where one of the most well known for finding the shortest path between two points in a graph is Dijkstra's algorithm.

To make sense of the graph data, there must be a way to express queries to extract some subset of the graph meeting given properties, such as the special case of finding a shortest path (a totally ordered sub-graph).
A response could be more general than that such as a DAG (where a number of directional trees and paths are special cases).
It should be noted that the underlying graph in many application cases can be structured as a DAG or special case thereof, meaning that the traversal would be uncomplicated.
As mentioned, one candidate for graph queries is GraphQL, where tree structures can effectively be extracted from the underlying general graph by providing query expressions for matching.

\subsubsection{Synthesizing a Data Model}

One of the key success factors in neuromorphic--digital integration is to find a data model in the NSP that is suitable for both the digital domain and the neuromorphic domain.
The objective is to have a data model in the NSP where data can be filled out by the neuromorphic system according to declarative instructions, and then effectively be provided through the digital interfaces.

The definition of the precise data model is highly application-dependent.
The objective in our integration framework is that there is a generic foundation providing graph-oriented data modeling, so there is a well-defined syntax to define the exact data model for each application case.
This application-specific model is set by the configuration interface of the NSP, and/or by software defining the application-specific NSP.
The same configuration information is also used in concert for SNN config.
This is to facilitate that the output spike-data from the NC system can be used by the NSP to fill out the data according to the model.
At certain moments, the NSP can trigger a digital event (e.g., caused directly by some specific NC output spike or that the data model is filled with data that should result in an event at the digital level).
This resembles an abstract thresholding for digital event triggering.
Subscribers of such events can react, and when receiving interesting events, they can access more information by placing a graph query to the NSP to fetch desired parts of the graph data.

Many of the challenges in this come from that state models of the two domains are fundamentally different, but they are also complementary:
\begin{enumerate}
    
    \item The neuromorphic state is continuously building up and decaying based on stimuli, which gives rise to spikes from some neurons that in turn stimulate other neurons, resulting in some transient state of the whole system.
The output spike data from the final layer of the neuromorphic system should be used to populate the data model of the NSP.
How this is to be accomplished is a subject for future research.

    \item In the digital domain, state is typically not decaying continuously.
On the contrary, in the NSP, it is possible to remember and store state over time (although at an abstract level compared to the neuromorphic system).
This is a complementary property to the neuromorphic state.
However, the NSPs are microservices typically operating close to their NC systems, in an edge cloud where storage space will be limited.
To mitigate this, snapshotting can be done only at interesting events, and soft state can optionally be used to forget and clean out state not triggered to be maintained (typically binary, not continuous), and interesting state can be offloaded asynchronously to central clouds.
    
\end{enumerate}

\subsection{Declarative Programming}
\label{sec:declarative}

This section discusses declarative programming as a goal-oriented approach to configuring NC systems.
Declarative programming is a programming paradigm in which the logic of computations is described without describing their control flow.
In a declarative language, what is described is the goal that a given program is intended to achieve (i.e., the desired outcome) rather than an explicit description of the sequence of computational primitives that the program is to carry out.
This is a suitable approach for NC systems, since there is no predefined control flow, and no desire to engineer traditional control flows.
The promise and rationale of declarative approaches is that the focus is on describing “what”, not “how”.
Advantages include clarity and unambiguity on the objectives, and within those objectives, opens for automation by AI of the dynamic properties and placements, and that correctness can be checked against the declarative objectives using a validation procedure.
The declarations are free from imperative details (the “how”), and side effects.

In deep learning, declarative frameworks \citep{molino2021declarative}, such as Ludwig \citep{molino2019ludwig}, aim to facilitate a new generation of systems that are more user-friendly by focusing on defining the data-schema and tasks, rather than low-level neural network information.
However, while ANNs and deep learning techniques are partly brain-inspired and offer a valuable starting point for SNN algorithm discovery, they represent only a subsection of the space of spike-based neural computation \citep{roy2019towards} that is available to neuromorphic hardware \citep{indiveri2019importance} and hybrid SNN--ANN models \citep{zhao2022framework}, see \textbf{Figure~\ref{fig:rep_space}} and \textbf{Table~\ref{tab:coding_schemes}}.
NC abstractions need to be co-designed \citep{schuman2022opportunities} with the digitization requirements and NC modules to provide a seamless edge-to-cloud integration and overall efficient hybrid AI solutions \citep{zhao2022framework} to the general computational problems defined in the use cases.
Furthermore, constraints associated with the interfaces between the declarative programming level and SNN/ANN descriptions based on existing software frameworks \citep{qu2022review} need to be considered.

Machine learning models, AI objectives, and NC-system configurations are naturally defined in declarative languages \citep{gould2022deep, schmitt2017neuromorphic}, while the details of “how” are subject to optimization, search, and plasticity.
Over the last years, there has generally been a shift in modeling preferences, from a focus on ANN- and SNN-centric modeling towards high-level interfaces supporting modern machine learning methods and tasks.
By declarative programming, the desired outcomes can be used to partly autogenerate and configure the NC modules (PyNN, process model, etc.), and to verify the compliance of system results to the desired outcomes.
The outcomes would depend both on the correctness of the declarative statements and on the data that is fed to the system.

The compliance of such declarative systems would not be totally deterministic as is typically the case in traditional declarative programs, in which the outcome can be verified logically and deterministically.
This poses some new challenges in conjunction with NC systems.
The outcomes must be validated statistically, given the datasets.
This leads to a need for verification and validation that resembles traditional software testing
However, here, in contrast, declarative statements have been used to both autogenerate settings and to verify whether the desired outcome---according to the declarative statements---have been met, with some statistical reliability (e.g., standard deviations) modulo the input data that has been used in the testing.
We propose to denote this as a \textit{validation procedure}, rather than a testing procedure, to emphasize the difference.
By that, we also avoid confusing the validation procedure with testing neuromorphic circuits, e.g., to detect manufacturing defects and faulty circuits, and runtime failures \citep{9000110, 9808431}.

\subsubsection{Test and Validation Methodology}

As mentioned above, we denote the verification, validation, and testing based on declarative statements and data fed to the system as a validation procedure.
This procedure aims to assess the statistical reliability of the system in meeting the declarative statements in the context of the structural and transient properties of the said system.
As discussed in Section \ref{sec:nc_states}, the structural properties are subject to direct manipulation by external configuration, optimization, and learning algorithms, which, we assume, are defined using declarative statements.
The transient properties, on the other hand, depend on the data presented to the system and its structural properties.
Note that the validation procedure is likely to mainly address a higher layer of abstraction than that of the transient SNN properties.
In addition, given the dependence on input data, the transient state of a system that can be observed through a validation procedure may exist only in direct connection to when the data is presented to the system, while structural states may be observable at any time.

The validation procedure should focus on the system's capacity to \textit{perform} the task learned, and as such it can also be useful in training protocols.
Given that the inherent states of an NC system are updated at different time-scales, we believe that a validation procedure should aim to examine the system focusing on relevant use-case-specific time-scales.
In this context, structural states should be examined to determine whether the NC system demonstrate the desired capacity to \textit{converge} to a solution or \textit{learn} a given type of task according to declarative statements, provided that input data contain the needed information.

\subsubsection{Process Calculus}
\label{sec:process_calculus}

Process calculus is a category of computer scientific approaches for the formal modeling of concurrent systems, which may serve as a base for declarative definitions of desired outcomes for networked NC systems.
A process calculus provides a framework for high-level description of interaction, communication, and synchronization between independent processes or agents.
Communication between processes is accomplished through channels, which can forward events that include some data. 
Many processes can send and/or receive from channels, so that complex distributed connectivity can be modeled, and there can be many processes on both the sending and receiving end of each channel.
Channels are first class objects, i.e., they can be passed around as data on other channels to achieve dynamic connectivity between processes.

In the context of NC systems, there are new challenges to using process calculus related to the non-deterministic stochastic behavior of such systems, and to the fact that system validation depends on both the declarations and the data.
Nevertheless, there exists work tying process calculus to spiking neural systems \citep{ciobanu2022process}, and Intel's NC software-framework Lava is inspired by the process calculus communicating sequential processes (CSP), see Section~\ref{sec:nc_software}. 
\section{Conclusion}

We have described the current landscape of NC technology, focusing on aspects expected to become increasingly relevant for large-scale adoption and integration into the present digital--computational infrastructure.
Based on this analysis, we have proposed a microservice-based framework for NC-system integration, consisting of a neuromorphic-system proxy providing virtualization and communication capabilities, in combination with a declarative programming approach offering engineering-process abstraction.
We have also presented well-known concepts in computer science that could be combined as a basis for the realization of the proposed framework, comprising the key components listed in the text below with some concluding remarks.
\textbf{Table~\ref{tab:alignment}} summarizes the alignment of the identified research directions and concepts with the integration challenges presented in Section~\ref{sec:introduction}.
The three research directions address the following identified knowledge gaps:
\begin{enumerate}

    \item \textbf{Neuromorphic-system proxy:}
A central question for the NSP is how to virtually represent the capabilities and states of physical NC systems, so that reliable microservices can be provided.
Such representations need to balance level of detail with computational expense.
For instance, a full simulation of the neuronal dynamics and event communication within an NC system may be motivated at times, while not feasible to be constantly maintained.

    \item \textbf{Communication and data models:}
In the proposed framework, NC--DC communication requires the development of an interparadigmatic data model that allows fetching relevant information from physical NC systems beyond the sparse event notifications and latent dynamic states.
This is closely tied to the challenge of developing transcoding principles and operational semantics for interfacing with the spike-based representations of NC systems.
Case studies focusing on optimization of spike encoders indicate that the best performing solutions are application-specific, which suggests that generic solutions require a combination of semantic technologies and machine learning--based optimization.

    \item \textbf{Declarative programming:}
A generic NC-programming framework requires appropriate programming abstractions, which enable engineers with commonly available competence to implement machine learning and AI solutions.
Declarative programming, focusing on “what” rather than “how”, is a natural approach to describe learning objectives and constraints.
There may also be a need for further developments of generalized system hierarchies, concepts of completeness, and analytical frameworks to facilitate hardware--software compatibility, programming flexibility, and development productivity.
Furthermore, a declarative framework presupposes methods for testing and validation.
For NC systems, this testing and validation would not be fully deterministic and logical, but would rather be closely interconnected with the core machine learning methodology and statistical assessment.

\end{enumerate}

\begin{landscape}
    \begin{table}[tb]
\caption{\textbf{Overview of research directions and concepts with relations to neuromorphic systems integration challenges.}
The symbol “$\otimes$” indicates that a concept has been connected to neuromorphic systems in the literature discussed in this article, while “$\times$” indicates that the connection is introduced here.
All connections indicated in this table are subjects for further research.
The relations between these concepts and conventional digital computing are described in Section~\ref{sec:framework}, but are not illustrated here.
        }
        \centering
        \begin{tabular}{l c c c c}
            \toprule
            \textbf{Research directions}    & \multicolumn{4}{c}{\textbf{Neuromorphic systems integration challenges}}      \\
            \textbf{and concepts}\\
            \cmidrule(r){2-5}
                                                    & Communication     & Virtualization    & Programming   & Testing and   \\
                                                    &                   &                   &               & validation    \\
            \midrule
\bfseries
            Neuromorphic-system proxy               & $\times$          & $\times$          & $\times$      & $\times$      \\
            \quad Neuromorphic-system simulation    &                   & $\otimes$         &               & $\otimes$     \\
            \quad Microservices                     & $\times$          & $\times$          &               &               \\
\bfseries
            Communication and data models           & $\times$          & $\times$          &               &               \\
            \quad Semantic technologies             & $\otimes$         &                   &               &               \\
            \quad Embeddings                        & $\otimes$         &                   &               &               \\
            \quad Soft state                        & $\times$          & $\times$          &               &               \\
            \quad PubSub messaging                  & $\times$          &                   &               &               \\
            \quad Graph data and querying           & $\times$          & $\times$          & $\times$      &               \\
\bfseries
            Declarative programming                 &                   &                   & $\otimes$     & $\otimes$     \\
            \quad Process calculus                  &                   &                   & $\otimes$     &               \\
            \quad Machine learning                  &                   &                   & $\otimes$     & $\otimes$     \\
            \quad Statistical evaluation            &                   &                   &               & $\otimes$     \\
            \bottomrule
        \end{tabular}
        \label{tab:alignment}
    \end{table}
\end{landscape}

In conclusion, there is a need for further research on interparadigmatic NC--DC communication models and virtualization to establish the transparency, reliability, and security that is typically required by large-scale distributed computing applications in cyber-physical systems and the industrial Internet.
Furthermore, research on NC programming abstractions and related protocols for training, validation, and testing are required to efficiently develop, integrate, and maintain hybrid neuromorphic--digital AI solutions in such large-scale distributed computing systems of systems.

\section*{Funding}

This work was partially funded by
The Kempe Foundations under Contract~JCK-1809,
the Arrowhead Tools project (ECSEL JU Grant No.~737 459),
the DAIS project (KDT JU Grant No.~101007273),
and the AI@Edge project (Horizon Europe Grant No.~101015922). 
\section*{Acknowledgments}

Some of the ideas presented in this manuscript were presented orally at the 2022 Workshop on Large-Scale Neuromorphic Systems Integration\footnotemark\ in Luleå, Sweden.
\footnotetext{https://www.ltu.se/research/subjects/Maskininlarning/Workshoppar/NSI-workshop?l=en} 
\bibliographystyle{Frontiers-Harvard}
\bibliography{references}

\begin{thebibliography}{97}
\providecommand{\natexlab}[1]{#1}
\expandafter\ifx\csname urlstyle\endcsname\relax
  \providecommand{\doi}[1]{doi:\discretionary{}{}{}#1}\else
  \providecommand{\doi}{doi:\discretionary{}{}{}\begingroup
  \urlstyle{rm}\Url}\fi
\providecommand{\selectlanguage}[1]{\relax}
\providecommand{\bibAnnoteFile}[1]{%
  \IfFileExists{#1}{\begin{quotation}\noindent\textsc{Key:} #1\\
  \textsc{Annotation:}\ \input{#1}\end{quotation}}{}}
\providecommand{\bibAnnote}[2]{%
  \begin{quotation}\noindent\textsc{Key:} #1\\
  \textsc{Annotation:}\ #2\end{quotation}}

\bibitem[{Aimone et~al.(2022)Aimone, Date, Fonseca-Guerra, Hamilton, Henke, Kay
  et~al.}]{aimone2022review}
Aimone, J., Date, P., Fonseca-Guerra, G., Hamilton, K., Henke, K., Kay, B.,
  et~al. (2022).
\newblock A review of non-cognitive applications for neuromorphic computing.
\newblock \emph{Neuromorphic Computing and Engineering}
\bibAnnoteFile{aimone2022review}

\bibitem[{Aimone et~al.(2019)Aimone, Severa, and Vineyard}]{aimone2019fugu}
Aimone, J.~B., Severa, W., and Vineyard, C.~M. (2019).
\newblock Composing neural algorithms with {Fugu}.
\newblock In \emph{Proceedings of the International Conference on Neuromorphic
  Systems}. 1--8
\bibAnnoteFile{aimone2019fugu}

\bibitem[{Astrom and Bernhardsson(2002)}]{astrom2002comparison}
Astrom, K. and Bernhardsson, B. (2002).
\newblock Comparison of {Riemann} and {Lebesgue} sampling for first order
  stochastic systems.
\newblock In \emph{Proceedings of the 41st IEEE Conference on Decision and
  Control, 2002.} vol.~2, 2011--2016 vol.2.
\newblock \doi{10.1109/CDC.2002.1184824}
\bibAnnoteFile{astrom2002comparison}

\bibitem[{Basu et~al.(2022)Basu, Deng, Frenkel, and Zhang}]{basu2022spiking}
Basu, A., Deng, L., Frenkel, C., and Zhang, X. (2022).
\newblock Spiking neural network integrated circuits: A review of trends and
  future directions.
\newblock In \emph{2022 IEEE Custom Integrated Circuits Conference (CICC)}.
  1--8.
\newblock \doi{10.1109/CICC53496.2022.9772783}
\bibAnnoteFile{basu2022spiking}

\bibitem[{Bauer et~al.(2022)Bauer, Lenz, Haghighatshoar, and
  Sheik}]{bauer2022exodus}
Bauer, F.~C., Lenz, G., Haghighatshoar, S., and Sheik, S. (2022).
\newblock {EXODUS}: Stable and efficient training of spiking neural networks.
\newblock \emph{arXiv preprint arXiv:2205.10242}
  \doi{10.48550/ARXIV.2205.10242}
\bibAnnoteFile{bauer2022exodus}

\bibitem[{Becker et~al.(2022)Becker, Haas, Schemmel, Furber, and
  Dolas}]{becker2022unconventional}
Becker, T., Haas, R., Schemmel, J., Furber, S., and Dolas, S. (2022).
\newblock \emph{Unconventional {HPC} Architectures}.
\newblock Tech. rep., European Technology Platform for High Performance
  Computing (ETP4HPC)
\bibAnnoteFile{becker2022unconventional}

\bibitem[{Bekolay et~al.(2014)Bekolay, Bergstra, Hunsberger, DeWolf, Stewart,
  Rasmussen et~al.}]{bekolay2014nengo}
Bekolay, T., Bergstra, J., Hunsberger, E., DeWolf, T., Stewart, T., Rasmussen,
  D., et~al. (2014).
\newblock Nengo: a {Python} tool for building large-scale functional brain
  models.
\newblock \emph{Frontiers in Neuroinformatics} 7.
\newblock \doi{10.3389/fninf.2013.00048}
\bibAnnoteFile{bekolay2014nengo}

\bibitem[{Boahen(2000)}]{boahen2000point}
Boahen, K. (2000).
\newblock Point-to-point connectivity between neuromorphic chips using address
  events.
\newblock \emph{IEEE Transactions on Circuits and Systems II: Analog and
  Digital Signal Processing} 47, 416--434.
\newblock \doi{10.1109/82.842110}
\bibAnnoteFile{boahen2000point}

\bibitem[{Brette(2015)}]{brette2015philosophy}
Brette, R. (2015).
\newblock Philosophy of the spike: Rate-based vs. spike-based theories of the
  brain.
\newblock \emph{Frontiers in Systems Neuroscience} 9.
\newblock \doi{10.3389/fnsys.2015.00151}
\bibAnnoteFile{brette2015philosophy}

\bibitem[{Chicca et~al.(2014)Chicca, Stefanini, Bartolozzi, and
  Indiveri}]{chicca2014neuromorphic}
Chicca, E., Stefanini, F., Bartolozzi, C., and Indiveri, G. (2014).
\newblock Neuromorphic electronic circuits for building autonomous cognitive
  systems.
\newblock \emph{Proceedings of the IEEE} 102, 1367--1388.
\newblock \doi{10.1109/JPROC.2014.2313954}
\bibAnnoteFile{chicca2014neuromorphic}

\bibitem[{Christensen et~al.(2022)Christensen, Dittmann, Linares-Barranco,
  Sebastian, Gallo, Redaelli et~al.}]{christensen2022roadmap}
Christensen, D.~V., Dittmann, R., Linares-Barranco, B., Sebastian, A., Gallo,
  M.~L., Redaelli, A., et~al. (2022).
\newblock 2022 roadmap on neuromorphic computing and engineering.
\newblock \emph{Neuromorphic Computing and Engineering} 2, 022501.
\newblock \doi{10.1088/2634-4386/ac4a83}
\bibAnnoteFile{christensen2022roadmap}

\bibitem[{Ciobanu and Todoran(2022)}]{ciobanu2022process}
Ciobanu, G. and Todoran, E.~N. (2022).
\newblock A process calculus for spiking neural {P} systems.
\newblock \emph{Information Sciences} 604, 298--319.
\newblock \doi{https://doi.org/10.1016/j.ins.2022.03.096}
\bibAnnoteFile{ciobanu2022process}

\bibitem[{Corradi and Indiveri(2015)}]{corradi2015neuromorphic}
Corradi, F. and Indiveri, G. (2015).
\newblock A neuromorphic event-based neural recording system for smart
  brain-machine-interfaces.
\newblock \emph{IEEE Transactions on Biomedical Circuits and Systems} 9,
  699--709.
\newblock \doi{10.1109/TBCAS.2015.2479256}
\bibAnnoteFile{corradi2015neuromorphic}

\bibitem[{Davari et~al.(1995)Davari, Dennard, and Shahidi}]{davari1995cmos}
Davari, B., Dennard, R., and Shahidi, G. (1995).
\newblock {CMOS} scaling for high performance and low power-the next ten years.
\newblock \emph{Proceedings of the IEEE} 83, 595--606.
\newblock \doi{10.1109/5.371968}
\bibAnnoteFile{davari1995cmos}

\bibitem[{Davies et~al.(2021)Davies, Wild, Orchard, Sandamirskaya, Guerra,
  Joshi et~al.}]{davies2021advancing}
Davies, M., Wild, A., Orchard, G., Sandamirskaya, Y., Guerra, G. A.~F., Joshi,
  P., et~al. (2021).
\newblock Advancing neuromorphic computing with {Loihi}: A survey of results
  and outlook.
\newblock \emph{Proceedings of the IEEE} 109, 911--934.
\newblock \doi{10.1109/JPROC.2021.3067593}
\bibAnnoteFile{davies2021advancing}

\bibitem[{Davison et~al.(2009)Davison, Br{\"u}derle, Eppler, Kremkow, Muller,
  Pecevski et~al.}]{davison2009pynn}
Davison, A.~P., Br{\"u}derle, D., Eppler, J.~M., Kremkow, J., Muller, E.,
  Pecevski, D., et~al. (2009).
\newblock {PyNN}: a common interface for neuronal network simulators.
\newblock \emph{Frontiers in neuroinformatics} 2, 11
\bibAnnoteFile{davison2009pynn}

\bibitem[{Di~Francesco(2017)}]{7958492}
Di~Francesco, P. (2017).
\newblock Architecting microservices.
\newblock In \emph{2017 IEEE International Conference on Software Architecture
  Workshops (ICSAW)}. 224--229.
\newblock \doi{10.1109/ICSAW.2017.65}
\bibAnnoteFile{7958492}

\bibitem[{Diamond et~al.(2016)Diamond, Nowotny, and
  Schmuker}]{diamond2016comparing}
Diamond, A., Nowotny, T., and Schmuker, M. (2016).
\newblock Comparing neuromorphic solutions in action: Implementing a
  bio-inspired solution to a benchmark classification task on three
  parallel-computing platforms.
\newblock \emph{Frontiers in Neuroscience} 9.
\newblock \doi{10.3389/fnins.2015.00491}
\bibAnnoteFile{diamond2016comparing}

\bibitem[{Dold et~al.(2022)Dold, Soler~Garrido, Caceres~Chian, Hildebrandt, and
  Runkler}]{dold2022neuro}
Dold, D., Soler~Garrido, J., Caceres~Chian, V., Hildebrandt, M., and Runkler,
  T. (2022).
\newblock Neuro-symbolic computing with spiking neural networks.
\newblock In \emph{Proceedings of the International Conference on Neuromorphic
  Systems 2022} (New York, NY, USA: Association for Computing Machinery), ICONS
  '22, 1--4.
\newblock \doi{10.1145/3546790.3546824}
\bibAnnoteFile{dold2022neuro}

\bibitem[{Eugster et~al.(2003)Eugster, Felber, Guerraoui, and
  Kermarrec}]{10.1145/857076.857078}
Eugster, P.~T., Felber, P.~A., Guerraoui, R., and Kermarrec, A.-M. (2003).
\newblock The many faces of publish/subscribe.
\newblock \emph{ACM Comput. Surv.} 35, 114–131.
\newblock \doi{10.1145/857076.857078}
\bibAnnoteFile{10.1145/857076.857078}

\bibitem[{Fall and Farrell(2008)}]{Fall2008}
Fall, K. and Farrell, S. (2008).
\newblock {DTN: An architectural retrospective}.
\newblock \emph{IEEE Journal on Selected Areas in Communications}
  \doi{10.1109/JSAC.2008.080609}
\bibAnnoteFile{Fall2008}

\bibitem[{Frenkel et~al.(2021)Frenkel, Bol, and Indiveri}]{frenkel2021bottom}
Frenkel, C., Bol, D., and Indiveri, G. (2021).
\newblock Bottom-up and top-down neural processing systems design: Neuromorphic
  intelligence as the convergence of natural and artificial intelligence.
\newblock \emph{arXiv preprint arXiv:2106.01288}
\bibAnnoteFile{frenkel2021bottom}

\bibitem[{Garcez and Lamb(2020)}]{garcez2020neurosymbolic}
Garcez, A.~d. and Lamb, L.~C. (2020).
\newblock Neurosymbolic {AI}: The 3rd wave.
\newblock \emph{arXiv preprint arXiv:2012.05876}
  \doi{10.48550/ARXIV.2012.05876}
\bibAnnoteFile{garcez2020neurosymbolic}

\bibitem[{Gebregiorgis and Tahoori(2019)}]{9000110}
Gebregiorgis, A. and Tahoori, M.~B. (2019).
\newblock Testing of neuromorphic circuits: Structural vs functional.
\newblock In \emph{2019 IEEE International Test Conference (ITC)}. 1--10.
\newblock \doi{10.1109/ITC44170.2019.9000110}
\bibAnnoteFile{9000110}

\bibitem[{Gewaltig and Diesmann(2007)}]{gewaltig2007nest}
Gewaltig, M.-O. and Diesmann, M. (2007).
\newblock {NEST (NEural Simulation Tool)}.
\newblock \emph{Scholarpedia} 2, 1430
\bibAnnoteFile{gewaltig2007nest}

\bibitem[{G{\"o}ltz et~al.(2021)G{\"o}ltz, Kriener, Baumbach, Billaudelle,
  Breitwieser, Cramer et~al.}]{goltz2021fast}
G{\"o}ltz, J., Kriener, L., Baumbach, A., Billaudelle, S., Breitwieser, O.,
  Cramer, B., et~al. (2021).
\newblock Fast and energy-efficient neuromorphic deep learning with first-spike
  times.
\newblock \emph{Nature Machine Intelligence} 3, 823--835.
\newblock \doi{10.1038/s42256-021-00388-x}
\bibAnnoteFile{goltz2021fast}

\bibitem[{Gould et~al.(2022)Gould, Hartley, and Campbell}]{gould2022deep}
Gould, S., Hartley, R., and Campbell, D. (2022).
\newblock Deep declarative networks.
\newblock \emph{IEEE Transactions on Pattern Analysis and Machine Intelligence}
  44, 3988--4004.
\newblock \doi{10.1109/TPAMI.2021.3059462}
\bibAnnoteFile{gould2022deep}

\bibitem[{Grasic et~al.(2011)Grasic, Davies, Lindgren, and Doria}]{Grasic2011}
Grasic, S., Davies, E., Lindgren, A., and Doria, A. (2011).
\newblock {The evolution of a DTN routing protocol - PRoPHETv2}.
\newblock In \emph{Proceedings of the Annual International Conference on Mobile
  Computing and Networking, MOBICOM}. 27--30.
\newblock \doi{10.1145/2030652.2030661}
\bibAnnoteFile{Grasic2011}

\bibitem[{Guo et~al.(2021{\natexlab{a}})Guo, Fouda, Eltawil, and
  Salama}]{guo2021neural}
Guo, W., Fouda, M.~E., Eltawil, A.~M., and Salama, K.~N. (2021{\natexlab{a}}).
\newblock Neural coding in spiking neural networks: A comparative study for
  robust neuromorphic systems.
\newblock \emph{Frontiers in Neuroscience} 15.
\newblock \doi{10.3389/fnins.2021.638474}
\bibAnnoteFile{guo2021neural}

\bibitem[{Guo et~al.(2021{\natexlab{b}})Guo, Zou, Hu, Yang, Wang, He
  et~al.}]{guo2021marr}
Guo, Y., Zou, X., Hu, Y., Yang, Y., Wang, X., He, Y., et~al.
  (2021{\natexlab{b}}).
\newblock A {Marr's} three-level analytical framework for neuromorphic
  electronic systems.
\newblock \emph{Advanced Intelligent Systems} , 2100054
\bibAnnoteFile{guo2021marr}

\bibitem[{Hamilton et~al.(2020)Hamilton, Schuman, Young, Bennink, Imam, and
  Humble}]{hamilton2020accelerating}
Hamilton, K.~E., Schuman, C.~D., Young, S.~R., Bennink, R.~S., Imam, N., and
  Humble, T.~S. (2020).
\newblock Accelerating scientific computing in the post-{Moore’s} era.
\newblock \emph{ACM Trans. Parallel Comput.} 7.
\newblock \doi{10.1145/3380940}
\bibAnnoteFile{hamilton2020accelerating}

\bibitem[{Hazan et~al.(2018)Hazan, Saunders, Khan, Patel, Sanghavi, Siegelmann
  et~al.}]{hazan2018bindsnet}
Hazan, H., Saunders, D.~J., Khan, H., Patel, D., Sanghavi, D.~T., Siegelmann,
  H.~T., et~al. (2018).
\newblock {BindsNET}: A machine learning-oriented spiking neural networks
  library in {Python}.
\newblock \emph{Frontiers in neuroinformatics} 12, 89
\bibAnnoteFile{hazan2018bindsnet}

\bibitem[{Hsieh et~al.(2021)Hsieh, Tseng, Chiu, and Li}]{9808431}
Hsieh, Y.-Z., Tseng, H.-Y., Chiu, I.-W., and Li, J. C.~M. (2021).
\newblock Fault modeling and testing of spiking neural network chips.
\newblock In \emph{2021 IEEE International Test Conference in Asia (ITC-Asia)}.
  1--6.
\newblock \doi{10.1109/ITC-Asia53059.2021.9808431}
\bibAnnoteFile{9808431}

\bibitem[{Indiveri and Liu(2015)}]{indiveri2015memory}
Indiveri, G. and Liu, S.-C. (2015).
\newblock Memory and information processing in neuromorphic systems.
\newblock \emph{Proceedings of the IEEE} 103, 1379--1397.
\newblock \doi{10.1109/JPROC.2015.2444094}
\bibAnnoteFile{indiveri2015memory}

\bibitem[{Indiveri and Sandamirskaya(2019)}]{indiveri2019importance}
Indiveri, G. and Sandamirskaya, Y. (2019).
\newblock The importance of space and time for signal processing in
  neuromorphic agents: The challenge of developing low-power, autonomous agents
  that interact with the environment.
\newblock \emph{IEEE Signal Processing Magazine} 36, 16--28.
\newblock \doi{10.1109/MSP.2019.2928376}
\bibAnnoteFile{indiveri2019importance}

\bibitem[{Jaeger(2021)}]{jaeger2021toward}
Jaeger, H. (2021).
\newblock Toward a generalized theory comprising digital, neuromorphic, and
  unconventional computing.
\newblock \emph{Neuromorphic Computing and Engineering}
\bibAnnoteFile{jaeger2021toward}

\bibitem[{Jaeger et~al.(2021)Jaeger, Doorakkers, Lawrence, and
  Indiveri}]{jaeger2021dimensions}
Jaeger, H., Doorakkers, D., Lawrence, C., and Indiveri, G. (2021).
\newblock Dimensions of timescales in neuromorphic computing systems.
\newblock \emph{arXiv preprint arXiv:2102.10648}
  \doi{10.48550/ARXIV.2102.10648}
\bibAnnoteFile{jaeger2021dimensions}

\bibitem[{Ji et~al.(2007)Ji, Ge, Kurose, and Towsley}]{Ji2007}
Ji, P., Ge, Z., Kurose, J., and Towsley, D. (2007).
\newblock {A comparison of hard-state and soft-state signaling protocols}.
\newblock \emph{IEEE/ACM Transactions on Networking}
  \doi{10.1109/TNET.2007.892849}
\bibAnnoteFile{Ji2007}

\bibitem[{Larrucea et~al.(2018)Larrucea, Santamaria, Colomo-Palacios, and
  Ebert}]{larrucea2018microservices}
Larrucea, X., Santamaria, I., Colomo-Palacios, R., and Ebert, C. (2018).
\newblock Microservices.
\newblock \emph{IEEE Software} 35, 96--100.
\newblock \doi{10.1109/MS.2018.2141030}
\bibAnnoteFile{larrucea2018microservices}

\bibitem[{Leiserson et~al.(2020)Leiserson, Thompson, Emer, Kuszmaul, Lampson,
  Sanchez et~al.}]{leiserson2020there}
Leiserson, C.~E., Thompson, N.~C., Emer, J.~S., Kuszmaul, B.~C., Lampson,
  B.~W., Sanchez, D., et~al. (2020).
\newblock There’s plenty of room at the top: What will drive computer
  performance after {Moore’s} law?
\newblock \emph{Science} 368, eaam9744.
\newblock \doi{10.1126/science.aam9744}
\bibAnnoteFile{leiserson2020there}

\bibitem[{Levina and Stantchev(2009)}]{5072495}
Levina, O. and Stantchev, V. (2009).
\newblock Realizing event-driven {SOA}.
\newblock In \emph{2009 Fourth International Conference on Internet and Web
  Applications and Services}. 37--42.
\newblock \doi{10.1109/ICIW.2009.14}
\bibAnnoteFile{5072495}

\bibitem[{Liu et~al.(2019)Liu, Rueckauer, Ceolini, Huber, and
  Delbruck}]{liu2019event}
Liu, S.-C., Rueckauer, B., Ceolini, E., Huber, A., and Delbruck, T. (2019).
\newblock Event-driven sensing for efficient perception: Vision and audition
  algorithms.
\newblock \emph{IEEE Signal Processing Magazine} 36, 29--37.
\newblock \doi{10.1109/MSP.2019.2928127}
\bibAnnoteFile{liu2019event}

\bibitem[{Liu et~al.(2003)Liu, Plale et~al.}]{liu2003survey}
Liu, Y., Plale, B., et~al. (2003).
\newblock Survey of publish subscribe event systems.
\newblock \emph{Computer Science Dept, Indian University} 16
\bibAnnoteFile{liu2003survey}

\bibitem[{Maass(1997)}]{maass1997networks}
Maass, W. (1997).
\newblock Networks of spiking neurons: The third generation of neural network
  models.
\newblock \emph{Neural Networks} 10, 1659--1671.
\newblock \doi{https://doi.org/10.1016/S0893-6080(97)00011-7}
\bibAnnoteFile{maass1997networks}

\bibitem[{Markovi{\'c} et~al.(2020)Markovi{\'c}, Mizrahi, Querlioz, and
  Grollier}]{markovic2020physics}
Markovi{\'c}, D., Mizrahi, A., Querlioz, D., and Grollier, J. (2020).
\newblock Physics for neuromorphic computing.
\newblock \emph{Nature Reviews Physics} 2, 499--510.
\newblock \doi{10.1038/s42254-020-0208-2}
\bibAnnoteFile{markovic2020physics}

\bibitem[{Mead(1990)}]{mead1990neuromorphic}
Mead, C. (1990).
\newblock Neuromorphic electronic systems.
\newblock \emph{Proceedings of the IEEE} 78, 1629--1636
\bibAnnoteFile{mead1990neuromorphic}

\bibitem[{Mead(2020)}]{mead2020we}
Mead, C. (2020).
\newblock How we created neuromorphic engineering.
\newblock \emph{Nature Electronics} 3, 434--435
\bibAnnoteFile{mead2020we}

\bibitem[{Mehonic and Kenyon(2022)}]{mehonic2022brain}
Mehonic, A. and Kenyon, A.~J. (2022).
\newblock Brain-inspired computing needs a master plan.
\newblock \emph{Nature} 604, 255--260
\bibAnnoteFile{mehonic2022brain}

\bibitem[{Mei et~al.(2022)Mei, Mao, Wang, Gan, and Tenenbaum}]{mei2022falcon}
Mei, L., Mao, J., Wang, Z., Gan, C., and Tenenbaum, J.~B. (2022).
\newblock Falcon: Fast visual concept learning by integrating images,
  linguistic descriptions, and conceptual relations.
\newblock \emph{arXiv preprint arXiv:2203.16639}
  \doi{10.48550/ARXIV.2203.16639}
\bibAnnoteFile{mei2022falcon}

\bibitem[{Molino et~al.(2019)Molino, Dudin, and Miryala}]{molino2019ludwig}
Molino, P., Dudin, Y., and Miryala, S.~S. (2019).
\newblock Ludwig: A type-based declarative deep learning toolbox.
\newblock \emph{arXiv preprint arXiv:1909.07930}
\bibAnnoteFile{molino2019ludwig}

\bibitem[{Molino and R\'{e}(2021)}]{molino2021declarative}
Molino, P. and R\'{e}, C. (2021).
\newblock Declarative machine learning systems.
\newblock \emph{Commun. ACM} 65, 42–49.
\newblock \doi{10.1145/3475167}
\bibAnnoteFile{molino2021declarative}

\bibitem[{Mortara and Vittoz(1994)}]{mortara1994communication}
Mortara, A. and Vittoz, E. (1994).
\newblock A communication architecture tailored for analog {VLSI} artificial
  neural networks: intrinsic performance and limitations.
\newblock \emph{IEEE Transactions on Neural Networks} 5, 459--466.
\newblock \doi{10.1109/72.286916}
\bibAnnoteFile{mortara1994communication}

\bibitem[{Müller et~al.(2022)Müller, Schmitt, Mauch, Billaudelle, Grübl,
  Güttler et~al.}]{muller2022brainscalesos}
Müller, E., Schmitt, S., Mauch, C., Billaudelle, S., Grübl, A., Güttler, M.,
  et~al. (2022).
\newblock The operating system of the neuromorphic {BrainScaleS-1} system.
\newblock \emph{Neurocomputing} 501, 790--810.
\newblock \doi{https://doi.org/10.1016/j.neucom.2022.05.081}
\bibAnnoteFile{muller2022brainscalesos}

\bibitem[{Nilsson(2022)}]{JNilsson2022PhDThesis}
Nilsson, J. (2022).
\newblock \emph{Machine Learning Concepts for Service Data Interoperability}.
\newblock Ph.D. thesis, Luleå University of Technology, Embedded Internet
  Systems Lab (EISLAB)
\bibAnnoteFile{JNilsson2022PhDThesis}

\bibitem[{{Nilsson} et~al.(2020){Nilsson}, {Delsing}, and
  {Sandin}}]{nilsson2020interoperability}
{Nilsson}, J., {Delsing}, J., and {Sandin}, F. (2020).
\newblock Autoencoder alignment approach to run-time interoperability for
  system of systems engineering.
\newblock In \emph{2020 IEEE 24th International Conference on Intelligent
  Engineering Systems (INES)}. 139--144
\bibAnnoteFile{nilsson2020interoperability}

\bibitem[{Nilsson and Sandin(2018)}]{nilsson2018semantic}
Nilsson, J. and Sandin, F. (2018).
\newblock Semantic interoperability in industry 4.0: Survey of recent
  developments and outlook.
\newblock In \emph{2018 IEEE 15th International Conference on Industrial
  Informatics (INDIN)} (IEEE), 127--132
\bibAnnoteFile{nilsson2018semantic}

\bibitem[{{Nilsson} et~al.(2019){Nilsson}, {Sandin}, and
  {Delsing}}]{nilsson2019interoperability}
{Nilsson}, J., {Sandin}, F., and {Delsing}, J. (2019).
\newblock Interoperability and machine-to-machine translation model with
  mappings to machine learning tasks.
\newblock In \emph{2019 IEEE 17th International Conference on Industrial
  Informatics (INDIN)}. vol.~1, 284--289
\bibAnnoteFile{nilsson2019interoperability}

\bibitem[{Nilsson(2021)}]{nilsson2021using}
Nilsson, M. (2021).
\newblock \emph{Using Inhomogeneous Neuronal--Synaptic Dynamics for
  Spatiotemporal Pattern Recognition in Neuromorphic Processors}.
\newblock Lic.\ thesis, Lule{\aa} University of Technology, Lule{\aa}, Sweden
\bibAnnoteFile{nilsson2021using}

\bibitem[{Nunes et~al.(2022)Nunes, Carvalho, Carneiro, and
  Cardoso}]{nunes2022spiking}
Nunes, J.~D., Carvalho, M., Carneiro, D., and Cardoso, J.~S. (2022).
\newblock Spiking neural networks: A survey.
\newblock \emph{IEEE Access} , 1--1\doi{10.1109/ACCESS.2022.3179968}
\bibAnnoteFile{nunes2022spiking}

\bibitem[{Perkins et~al.(2003)Perkins, Belding-Royer, and Das}]{Perkins2003}
Perkins, C., Belding-Royer, E., and Das, S. (2003).
\newblock {RFC 3561-ad hoc on-demand distance vector (AODV) routing}.
\newblock \emph{Internet RFCs}
\bibAnnoteFile{Perkins2003}

\bibitem[{Perrey and Lycett(2003)}]{1210138}
Perrey, R. and Lycett, M. (2003).
\newblock Service-oriented architecture.
\newblock In \emph{2003 Symposium on Applications and the Internet Workshops,
  2003. Proceedings.} 116--119.
\newblock \doi{10.1109/SAINTW.2003.1210138}
\bibAnnoteFile{1210138}

\bibitem[{Qu et~al.(2022)Qu, Yang, Zheng, and Zhang}]{qu2022review}
Qu, P., Yang, L., Zheng, W., and Zhang, Y. (2022).
\newblock A review of basic software for brain-inspired computing.
\newblock \emph{CCF Transactions on High Performance Computing} , 1--9
\bibAnnoteFile{qu2022review}

\bibitem[{Rao et~al.(2022)Rao, Plank, Wild, and Maass}]{rao2022long}
Rao, A., Plank, P., Wild, A., and Maass, W. (2022).
\newblock A long short-term memory for {AI} applications in spike-based
  neuromorphic hardware.
\newblock \emph{Nature Machine Intelligence} , 1--13
\bibAnnoteFile{rao2022long}

\bibitem[{Roy et~al.(2019)Roy, Jaiswal, and Panda}]{roy2019towards}
Roy, K., Jaiswal, A., and Panda, P. (2019).
\newblock Towards spike-based machine intelligence with neuromorphic computing.
\newblock \emph{Nature} 575, 607--617
\bibAnnoteFile{roy2019towards}

\bibitem[{Rubino et~al.(2021)Rubino, Livanelioglu, Qiao, Payvand, and
  Indiveri}]{rubino2021ultra}
Rubino, A., Livanelioglu, C., Qiao, N., Payvand, M., and Indiveri, G. (2021).
\newblock Ultra-low-power {FDSOI} neural circuits for extreme-edge neuromorphic
  intelligence.
\newblock \emph{IEEE Transactions on Circuits and Systems I: Regular Papers}
  68, 45--56.
\newblock \doi{10.1109/TCSI.2020.3035575}
\bibAnnoteFile{rubino2021ultra}

\bibitem[{Rueckauer et~al.(2017)Rueckauer, Lungu, Hu, Pfeiffer, and
  Liu}]{rueckauer2017conversion}
Rueckauer, B., Lungu, I.-A., Hu, Y., Pfeiffer, M., and Liu, S.-C. (2017).
\newblock Conversion of continuous-valued deep networks to efficient
  event-driven networks for image classification.
\newblock \emph{Frontiers in Neuroscience} 11, 682.
\newblock \doi{10.3389/fnins.2017.00682}
\bibAnnoteFile{rueckauer2017conversion}

\bibitem[{Sandamirskaya(2014)}]{sandamirskaya2014dynamic}
Sandamirskaya, Y. (2014).
\newblock Dynamic neural fields as a step toward cognitive neuromorphic
  architectures.
\newblock \emph{Frontiers in Neuroscience} 7.
\newblock \doi{10.3389/fnins.2013.00276}
\bibAnnoteFile{sandamirskaya2014dynamic}

\bibitem[{Schmitt et~al.(2017)Schmitt, Klähn, Bellec, Grübl, Güttler, Hartel
  et~al.}]{schmitt2017neuromorphic}
Schmitt, S., Klähn, J., Bellec, G., Grübl, A., Güttler, M., Hartel, A.,
  et~al. (2017).
\newblock Neuromorphic hardware in the loop: Training a deep spiking network on
  the {BrainScaleS} wafer-scale system.
\newblock In \emph{2017 International Joint Conference on Neural Networks
  (IJCNN)}. 2227--2234.
\newblock \doi{10.1109/IJCNN.2017.7966125}
\bibAnnoteFile{schmitt2017neuromorphic}

\bibitem[{Schuman et~al.(2022{\natexlab{a}})Schuman, Rizzo, McDonald-Carmack,
  Skuda, and Plank}]{schuman2022evaluating}
Schuman, C., Rizzo, C., McDonald-Carmack, J., Skuda, N., and Plank, J.
  (2022{\natexlab{a}}).
\newblock Evaluating encoding and decoding approaches for spiking neuromorphic
  systems.
\newblock In \emph{Proceedings of the International Conference on Neuromorphic
  Systems 2022} (New York, NY, USA: Association for Computing Machinery), ICONS
  '22, 1--9.
\newblock \doi{10.1145/3546790.3546792}
\bibAnnoteFile{schuman2022evaluating}

\bibitem[{Schuman et~al.(2022{\natexlab{b}})Schuman, Kulkarni, Parsa, Mitchell,
  Date, and Kay}]{schuman2022opportunities}
Schuman, C.~D., Kulkarni, S.~R., Parsa, M., Mitchell, J.~P., Date, P., and Kay,
  B. (2022{\natexlab{b}}).
\newblock Opportunities for neuromorphic computing algorithms and applications.
\newblock \emph{Nature Computational Science} 2, 10--19.
\newblock \doi{10.1038/s43588-021-00184-y}
\bibAnnoteFile{schuman2022opportunities}

\bibitem[{Schuman et~al.(2020)Schuman, Mitchell, Patton, Potok, and
  Plank}]{schuman2020evolutionary}
Schuman, C.~D., Mitchell, J.~P., Patton, R.~M., Potok, T.~E., and Plank, J.~S.
  (2020).
\newblock Evolutionary optimization for neuromorphic systems.
\newblock In \emph{Proceedings of the Neuro-inspired Computational Elements
  Workshop}. 1--9
\bibAnnoteFile{schuman2020evolutionary}

\bibitem[{Severa et~al.(2019)Severa, Vineyard, Dellana, Verzi, and
  Aimone}]{severa2019whetstone}
Severa, W., Vineyard, C.~M., Dellana, R., Verzi, S.~J., and Aimone, J.~B.
  (2019).
\newblock Training deep neural networks for binary communication with the
  {Whetstone} method.
\newblock \emph{Nature Machine Intelligence} 1, 86--94
\bibAnnoteFile{severa2019whetstone}

\bibitem[{Shalf(2020)}]{shalf2020future}
Shalf, J. (2020).
\newblock The future of computing beyond {Moore’s} law.
\newblock \emph{Philosophical Transactions of the Royal Society A} 378,
  20190061
\bibAnnoteFile{shalf2020future}

\bibitem[{Shen(2010)}]{Shen2010}
Shen, H. (2010).
\newblock Content-based publish/subscribe systems.
\newblock In \emph{Handbook of Peer-to-Peer Networking}, eds. X.~Shen, H.~Yu,
  J.~Buford, and M.~Akon (Boston, MA: Springer US). 1333--1366.
\newblock \doi{10.1007/978-0-387-09751-0_49}
\bibAnnoteFile{Shen2010}

\bibitem[{Shrestha et~al.(2022)Shrestha, Fang, Mei, Rider, Wu, and
  Qiu}]{shrestha2022survey}
Shrestha, A., Fang, H., Mei, Z., Rider, D.~P., Wu, Q., and Qiu, Q. (2022).
\newblock A survey on neuromorphic computing: Models and hardware.
\newblock \emph{IEEE Circuits and Systems Magazine} 22, 6--35.
\newblock \doi{10.1109/MCAS.2022.3166331}
\bibAnnoteFile{shrestha2022survey}

\bibitem[{Shrestha and Orchard(2018)}]{shrestha2018slayer}
Shrestha, S.~B. and Orchard, G. (2018).
\newblock Slayer: Spike layer error reassignment in time.
\newblock \emph{arXiv preprint arXiv:1810.08646}
\bibAnnoteFile{shrestha2018slayer}

\bibitem[{Stewart(2012)}]{stewart2012technical}
Stewart, T.~C. (2012).
\newblock A technical overview of the neural engineering framework.
\newblock \emph{University of Waterloo} 110
\bibAnnoteFile{stewart2012technical}

\bibitem[{Stimberg et~al.(2019)Stimberg, Brette, and
  Goodman}]{stimberg2019brian2}
Stimberg, M., Brette, R., and Goodman, D.~F. (2019).
\newblock Brian 2, an intuitive and efficient neural simulator.
\newblock \emph{Elife} 8, e47314
\bibAnnoteFile{stimberg2019brian2}

\bibitem[{Stöckl and Maass(2021)}]{stockl2021optimized}
Stöckl, C. and Maass, W. (2021).
\newblock Optimized spiking neurons can classify images with high accuracy
  through temporal coding with two spikes.
\newblock \emph{Nature Machine Intelligence} 3, 230--238.
\newblock \doi{10.1038/s42256-021-00311-4}
\bibAnnoteFile{stockl2021optimized}

\bibitem[{Tayarani-Najaran and Schmuker(2021)}]{tayarani2021event}
Tayarani-Najaran, M.-H. and Schmuker, M. (2021).
\newblock Event-based sensing and signal processing in the visual, auditory,
  and olfactory domain: A review.
\newblock \emph{Frontiers in Neural Circuits} 15.
\newblock \doi{10.3389/fncir.2021.610446}
\bibAnnoteFile{tayarani2021event}

\bibitem[{Thompson et~al.(2021)Thompson, Greenewald, Lee, and
  Manso}]{thompson2021deep}
Thompson, N.~C., Greenewald, K., Lee, K., and Manso, G.~F. (2021).
\newblock Deep learning's diminishing returns: The cost of improvement is
  becoming unsustainable.
\newblock \emph{IEEE Spectrum} 58, 50--55.
\newblock \doi{10.1109/MSPEC.2021.9563954}
\bibAnnoteFile{thompson2021deep}

\bibitem[{Thorpe et~al.(2001)Thorpe, Delorme, and {Van
  Rullen}}]{thorpe2001spike}
Thorpe, S., Delorme, A., and {Van Rullen}, R. (2001).
\newblock Spike-based strategies for rapid processing.
\newblock \emph{Neural Networks} 14, 715--725.
\newblock \doi{https://doi.org/10.1016/S0893-6080(01)00083-1}
\bibAnnoteFile{thorpe2001spike}

\bibitem[{Vetter et~al.(2018)Vetter, Brightwell, Gokhale, McCormick, Ross,
  Shalf et~al.}]{vetter2018extreme}
Vetter, J.~S., Brightwell, R., Gokhale, M., McCormick, P., Ross, R., Shalf, J.,
  et~al. (2018).
\newblock \emph{Extreme Heterogeneity 2018 – Productive Computational Science
  in the Era of Extreme Heterogeneity: Report for {DOE ASCR Workshop on Extreme
  Heterogeneity}}.
\newblock Tech. rep., USDOE Office of Science (SC), Advanced Scientific
  Computing Research (ASCR).
\newblock \doi{10.2172/1473756}
\bibAnnoteFile{vetter2018extreme}

\bibitem[{Waldrop(2016)}]{waldrop2016chips}
Waldrop, M.~M. (2016).
\newblock The chips are down for {Moore’s} law.
\newblock \emph{Nature News} 530, 144
\bibAnnoteFile{waldrop2016chips}

\bibitem[{Wang et~al.(2022)Wang, Cheng, and Lim}]{wang2022hierarchical}
Wang, S., Cheng, T.~H., and Lim, M.~H. (2022).
\newblock A hierarchical taxonomic survey of spiking neural networks.
\newblock \emph{Memetic Computing} \doi{10.1007/s12293-022-00373-w}
\bibAnnoteFile{wang2022hierarchical}

\bibitem[{Wilson(2010)}]{wilson10}
Wilson, R.~J. (2010).
\newblock \emph{Introduction to Graph Theory} (New York: Prentice Hall/Pearson)
\bibAnnoteFile{wilson10}

\bibitem[{Xiao et~al.(2016)Xiao, Wijegunaratne, and Qiang}]{7880473}
Xiao, Z., Wijegunaratne, I., and Qiang, X. (2016).
\newblock Reflections on {SOA} and microservices.
\newblock In \emph{2016 4th International Conference on Enterprise Systems
  (ES)}. 60--67.
\newblock \doi{10.1109/ES.2016.14}
\bibAnnoteFile{7880473}

\bibitem[{Yavuz et~al.(2016)Yavuz, Turner, and Nowotny}]{yavuz2016genn}
Yavuz, E., Turner, J., and Nowotny, T. (2016).
\newblock {GeNN}: A code generation framework for accelerated brain
  simulations.
\newblock \emph{Scientific reports} 6, 1--14
\bibAnnoteFile{yavuz2016genn}

\bibitem[{Ye et~al.(2021)Ye, Wang, Liu, Chen, Li, Zhang
  et~al.}]{ye2021challenges}
Ye, L., Wang, Z., Liu, Y., Chen, P., Li, H., Zhang, H., et~al. (2021).
\newblock The challenges and emerging technologies for low-power artificial
  intelligence {IoT} systems.
\newblock \emph{IEEE Transactions on Circuits and Systems I: Regular Papers}
  68, 4821--4834.
\newblock \doi{10.1109/TCSI.2021.3095622}
\bibAnnoteFile{ye2021challenges}

\bibitem[{Yin et~al.(2021)Yin, Corradi, and Boht{\'e}}]{yin2021accurate}
Yin, B., Corradi, F., and Boht{\'e}, S.~M. (2021).
\newblock Accurate and efficient time-domain classification with adaptive
  spiking recurrent neural networks.
\newblock \emph{Nature Machine Intelligence} 3, 905--913
\bibAnnoteFile{yin2021accurate}

\bibitem[{Zenke et~al.(2021)Zenke, Bohté, Clopath, Comşa, Göltz, Maass
  et~al.}]{zenke2021visualizing}
Zenke, F., Bohté, S.~M., Clopath, C., Comşa, I.~M., Göltz, J., Maass, W.,
  et~al. (2021).
\newblock Visualizing a joint future of neuroscience and neuromorphic
  engineering.
\newblock \emph{Neuron} 109, 571--575.
\newblock \doi{https://doi.org/10.1016/j.neuron.2021.01.009}
\bibAnnoteFile{zenke2021visualizing}

\bibitem[{Zhang and Hu(2005)}]{1437105}
Zhang, R. and Hu, Y. (2005).
\newblock {HYPER}: A hybrid approach to efficient content-based
  publish/subscribe.
\newblock In \emph{25th IEEE International Conference on Distributed Computing
  Systems (ICDCS'05)}. 427--436.
\newblock \doi{10.1109/ICDCS.2005.42}
\bibAnnoteFile{1437105}

\bibitem[{Zhang et~al.(2020{\natexlab{a}})Zhang, Gao, Tang, Yao, Yu, Chang
  et~al.}]{zhang2020neuro}
Zhang, W., Gao, B., Tang, J., Yao, P., Yu, S., Chang, M.-F., et~al.
  (2020{\natexlab{a}}).
\newblock Neuro-inspired computing chips.
\newblock \emph{Nature electronics} 3, 371--382
\bibAnnoteFile{zhang2020neuro}

\bibitem[{Zhang et~al.(2020{\natexlab{b}})Zhang, Qu, Ji, Zhang, Gao, Wang
  et~al.}]{zhang2020system}
Zhang, Y., Qu, P., Ji, Y., Zhang, W., Gao, G., Wang, G., et~al.
  (2020{\natexlab{b}}).
\newblock A system hierarchy for brain-inspired computing.
\newblock \emph{Nature} 586, 378--384
\bibAnnoteFile{zhang2020system}

\bibitem[{Zhao et~al.(2022)Zhao, Yang, Zheng, Wu, Liu, Wu
  et~al.}]{zhao2022framework}
Zhao, R., Yang, Z., Zheng, H., Wu, Y., Liu, F., Wu, Z., et~al. (2022).
\newblock A framework for the general design and computation of hybrid neural
  networks.
\newblock \emph{Nature Communications} 13.
\newblock \doi{10.1038/s41467-022-30964-7}
\bibAnnoteFile{zhao2022framework}

\bibitem[{Zhou et~al.(2019)Zhou, Chen, Li, Zeng, Luo, and Zhang}]{zhou2019edge}
Zhou, Z., Chen, X., Li, E., Zeng, L., Luo, K., and Zhang, J. (2019).
\newblock Edge intelligence: Paving the last mile of artificial intelligence
  with edge computing.
\newblock \emph{Proceedings of the IEEE} 107, 1738--1762.
\newblock \doi{10.1109/JPROC.2019.2918951}
\bibAnnoteFile{zhou2019edge}

\bibitem[{Zidan et~al.(2018)Zidan, Strachan, and Lu}]{zidan2018future}
Zidan, M.~A., Strachan, J.~P., and Lu, W.~D. (2018).
\newblock The future of electronics based on memristive systems.
\newblock \emph{Nature electronics} 1, 22--29
\bibAnnoteFile{zidan2018future}

\end{thebibliography}

\end{document}